\documentclass[letterpaper, 10 pt, conference]{ieeeconf}  

\IEEEoverridecommandlockouts                              

\usepackage{graphicx}
\usepackage{amsmath}
\usepackage[backend=bibtex,style=ieee,natbib=true]{biblatex}

\usepackage[colorlinks=true,bookmarks=true]{hyperref}
\usepackage{afterpage}
\usepackage{pifont}
\usepackage[table]{xcolor}

\usepackage{xcolor}
\usepackage[small]{caption}

\usepackage{array}
\usepackage{booktabs}
\usepackage{multirow}

\usepackage{float}
\usepackage{soul}
\sethlcolor{red}

\usepackage{svg}

\usepackage{amssymb}

\title{\LARGE \bf
GenDOM: Generalizable One-shot Deformable Object
Manipulation \\
with Parameter-Aware Policy
}

\author{So Kuroki$^{1}$, Jiaxian Guo$^{1}$, Tatsuya Matsushima$^{1}$, Takuya Okubo$^{1}$,
Masato Kobayashi$^{2}$, Yuya Ikeda$^{1}$, \\
Ryosuke Takanami$^{1}$, Paul Yoo$^{1}$,Yutaka Matsuo$^{1}$, Yusuke Iwasawa$^{1}$
\thanks{$^{1}$The University of Tokyo, Japan}%
\thanks{$^{2}$Osaka University, Japan}%
\thanks{Email: so.kuroki1931@gmail.com}%
}

\begin{document}

\maketitle
\thispagestyle{empty}
\pagestyle{empty}

\begin{abstract}
Due to the inherent uncertainty in their deformability during motion, previous methods in deformable object manipulation, such as rope and cloth, often required hundreds of real-world demonstrations to train a manipulation policy for each object, which hinders their applications in our ever-changing world. 
To address this issue, we introduce \emph{GenDOM}, a framework that allows the manipulation policy to handle different deformable objects with only a single real-world demonstration. 
To achieve this, we augment the policy by conditioning it on deformable object parameters and training it with a diverse range of simulated deformable objects so that the policy can adjust actions based on different object parameters.
At the time of inference, given a new object, \emph{GenDOM} can estimate the deformable object parameters with only a single real-world demonstration by minimizing the disparity between the grid density of point clouds of real-world demonstrations and simulations in a differentiable physics simulator. 
Empirical validations on both simulated and real-world object manipulation setups clearly show that our method can manipulate different objects with a single demonstration and significantly outperforms the baseline in both environments (a 62\% improvement for in-domain ropes and a 15\% improvement for out-of-distribution ropes in simulation, as well as a 26\% improvement for ropes and a 50\% improvement for cloths in the real world), demonstrating the effectiveness of our approach in one-shot deformable object manipulation. \newline
\url{https://sites.google.com/view/gendom/home}.

\end{abstract}

\section{Introduction}
\label{sec:introduction}
\begin{figure*}[t]
    \begin{center}
    \includegraphics[width=\textwidth]{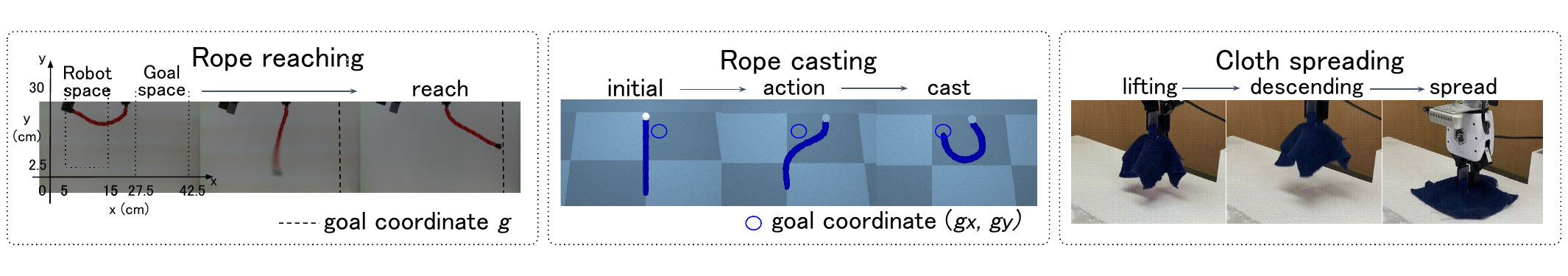}
    \vskip -0.2in
    \caption{Task visualization. \textbf{Rope reaching:} A robot arm guides a rope end towards a release coordinate, after which the rope moves gravitationally towards goal coordinate $g$. \textbf{Rope casting:} A robot arm performs a momentary action in mid-air, guiding the rope towards coordinate ($g_x, g_y$). \textbf{Cloth spreading:} A robot arm handles a cloth's center and acts with acceleration to spread it over a table.}
    \label{fig:tasks}
    \end{center}
    \vskip -0.3in
    
\end{figure*}


Deformable object manipulation constitutes a salient aspect of numerous routine tasks, encompassing domestic chores such as folding laundry, culinary endeavors, and the weaving of rope~\cite{dom-survey2, dom-survey3}. 
Despite the promising potential, learning a policy for it is still challenging. 
It often necessitates numerous real-world demonstrations due to the complex dynamics of deformable objects, which encompass unpredictable deformability, elasticity, and plasticity during motion~\cite{dom-survey2, dom-survey3}.
Nonetheless, the extensive volume of data required to train these policies, typically gathered through real robots, can often be costly and impractical~\cite{wang2014learning, nair2017combining}.


Recent research \cite{mehta2020active, rabinovitz2021unsupervised, leibovich2022validate, james2019sim,nachum2019multi} tried to alleviate this problem via Simulation-to-Real (Sim2Real) transfer, wherein policies are trained within simulation environments with unlimited data and subsequently applied to real-world tasks~\cite{matas2018sim}. 
Despite these efforts, mitigating Sim2Real gap continues to represent a protracted and complex problem within the robotics field~\cite{mehta2020active, rabinovitz2021unsupervised, leibovich2022validate, james2019sim,nachum2019multi,abeyruwan2023sim2real,pashevich2019learning,jakobi1995noise,mahler2017dex}. 
This challenge is exacerbated, especially when dealing with deformable objects situated in unpredictable dynamic environments. 
A common approach to these complexities is to train a distinct policy for each object, each possessing unique dynamics~\cite{lim2022real2sim2real, lee2022cavfm,angelina2019vpa}. This approach usually requires hundreds of real-world demonstrations for each object, which are both time-consuming and costly to collect.
This requirement for real-world data thus becomes a significant bottleneck for the scalability of robotics tasks, restricting the potential for broader application~\cite{lee2022cavfm,angelina2019vpa}.

In order to bridge the Sim2Real gap in deformable object manipulation, prior approaches such as Real2Sim2Real \cite{wang2022real2sim2real} are proposed, which initially carry out simulation identification to align the simulation with real-world dynamics, a process known as Real-to-Simulation (Real2Sim). Subsequently, they leverage these aligned simulations to assist in policy learning, thereby reducing the need for extensive real-world data~\cite{wang2022real2sim2real, wang2022real2sim2real2, lim2022real2sim2real}. Despite these advances, they still require a considerable volume of real-world demonstrations for effective simulation identification. Moreover, they are typically specialized for certain objects or tasks, which limits their generalizability to a broader spectrum of applications.

In this paper, we propose a framework \emph{GenDOM}, which allows the trained policy to manipulate different objects—even those with out-of-distribution dynamics—using just a single real-world demonstration, \emph{i.e.}, one-shot generalization~\cite{rezende2016one}. 
Our key idea is to condition the policy with specific deformable object parameters that can affect dynamics. 
Specifically, we select Young’s modulus and Poisson’s ratio~\cite{sanchez2018robotic, sengupta2020simultaneous, dom-pt1}
which are commonly utilized in modelling the deformation~\cite{sanchez2018robotic,sengupta2020simultaneous, dom-pt1}.
While these parameters hold significant potential in determining an object's deformability, their usage in policy training remains largely uncharted territory due to the inherent difficulties in estimating these values.
Our research represents an initial effort to integrate these parameters into manipulation policy training, thereby engendering a policy that is not only aware of object-specific parameters but is also attuned to dynamics. 
As a result, the policy can adjust its action output for various deformable objects.
To train such an augmented manipulation policy, we randomly sample different object parameters in the simulation and add them as the input of the policy network. 
This inclusion allows the policy to encompass a broader range of information on various dynamics parameters, utilizing a single network for deformable object manipulation (Figure~\ref{fig:tasks}).

Furthermore, we introduce a novel, efficient gradient-based Real2Sim technique specifically crafted for diverse deformable objects within a unified setup.
Leveraging the capabilities of differentiable physics, we propose a gradient-based technique \cite{sundaresan2022diffcloud, ma2022risp, le2023differentiable,huang2021plasticinelab, jatavallabhula2021gradsim, chendaxbench, heiden2021disect} to estimate the object parameters with only a single real-world demonstration. 
Our approach seeks to minimize the discrepancy of point clouds between the simulation and real-world environments. 
Contrary to prior Real2Sim2Real methodologies that utilized estimated deformable object parameters as simulation environmental parameters to learn the policy~\cite{lim2022real2sim2real}, our method uses these estimated parameters directly as the condition of the policy, enabling it to generalize to objects with differing dynamics.

In summary, our contributions are as follows: (1) We propose a policy augmented with parameters that account for deformable object characteristics (Young’s Modulus and Poisson’s ratio). We show that these parameters are effective in controlling deformable objects such as rope and cloth, and these can be estimated via designed gradient-based optimization given only a single real-world demonstration. (2) Empirical validation on both in-domain and out-of-distribution parameters of the dynamics demonstrate that GenDOM excels in terms of its generalization capability across dynamics, even with only a single real-world demonstration. Specifically, GenDOM provides a 62\% improvement for in-domain ropes and a 15\% improvement for out-of-distribution ropes in simulation, as well as a 26\% improvement for ropes and a 50\% improvement for cloths in the real world, highlighting its superior adaptability and potential for broad applicability.

\section{Related Work}
\label{sec:related_work}

\subsection{Deformable Object Manipulation}

Deformable object manipulation remains a challenging research area due to its inherent complexity. Each object possesses unique dynamics and specific degrees of freedom that demand careful consideration, differing notably from their rigid counterparts\cite{sanchez2018robotic, dom-survey2,dom-transporter-network}. 
Previous research has succeeded in manipulating various types of deformable objects, such as rope\cite{dom-rope}, cloth\cite{dom-cloth}, elasto-plastic\cite{dom-elasto}, and liquid\cite{dom-liquid}. 
Despite this progress, these methods are plagued with significant limitations.
The primary concerns are their substantial reliance on a large number of real-world demonstrations for effective learning and generalization capabilities for different scenarios~\cite{sanchez2018robotic, dom-survey2, dom-survey3, lee2022cavfm,angelina2019vpa}.
Various learning methods have been proposed to operate deformable objects with limited data~\cite{sundaresan2020learning, finn2017one}.
In this study, our method offers extensive generalization capabilities and benefits from one-shot learning, significantly reducing the dependency on numerous real-world demonstrations. It also shows comparable performance to policies trained with over 100 real-world datasets.




\subsection{Sim2Real}
Numerous studies focus on filling the gap between simulation and real-world, domain randomization~\cite{mehta2020active, rabinovitz2021unsupervised, leibovich2022validate, james2019sim}, hierarchical learning~\cite{nachum2019multi, abeyruwan2023sim2real}, and data augmentation~\cite{pashevich2019learning}. In pursuit of dynamics generalization,~\cite{murooka2021exi} integrated both explicit and implicit dynamic parameters into their policy in a rigid-object push manipulation task. In the realm of deformable object manipulation,~\cite{kuroki2023collective, qi2022learning, lin2022planning} leverage differential physics to collect demonstrations on deformable object manipulation and subsequently train a policy through imitation learning to Sim2Real. Nonetheless, these models for deformable object manipulation have limitations of generalization ability for different objects due to their difficulty in learning complex dynamics. To achieve this, we use parameters that account for deformable object characteristics to train our policy.

\subsection{Real2Sim}
The demand for collecting realistic deformable object data in simulation scenarios is escalating. 
Much research is limited due to using specific devices or sensors~\cite{dom-pt1, dom-pt2}.
While learning-based methods to estimate parameters of deformable objects in simulation have been proposed~\cite{yang2017learning}, these methods rely on huge datasets.
Thanks to the significant improvements in differentiable simulation~\cite{huang2021plasticinelab, jatavallabhula2021gradsim, chendaxbench, heiden2021disect}, several gradient-based methods have been introduced.
For example, \cite{sundaresan2022diffcloud} proposes differentiable point cloud sampling from mesh states.
\cite{ma2022risp} employed differentiable rendering methods. 
In this study, we aim to estimate deformable object parameters.
To ensure that the parameter captures the physical characteristics of objects, we select Young's modulus and Poisson's ratio to estimate, which is commonly used for modeling deformation as a linear and isotropic and homogeneous deformation model~\cite{sanchez2018robotic, sengupta2020simultaneous, dom-pt1}. 

\subsection{Real2Sim2Real}
Some research has been undertaken as Real2Sim2Real. \cite{wang2022real2sim2real} reconstructs high-quality meshes in simulation from real-world point clouds to achieve a robust policy in the real world.
\cite{wang2022real2sim2real2} bridges the Sim2Real gap by optimizing parameters for tensegrity robots on a simulation platform using real data via gradient descent.
\cite{lim2022real2sim2real} estimates parameters with Differential Evolution and executes a Sim2Real transfer with a policy trained by the dataset collected with the estimated parameters. While the preceding research advocates for a Real2Sim2Real pipeline, they focus on dealing with specific dynamics, and few studies pursue the generalization ability for several dynamics.

\section{Proposed Method}
\label{sec:proposed_method}
\begin{figure}[t]
    \begin{center}
    \includegraphics[width=0.48\textwidth]{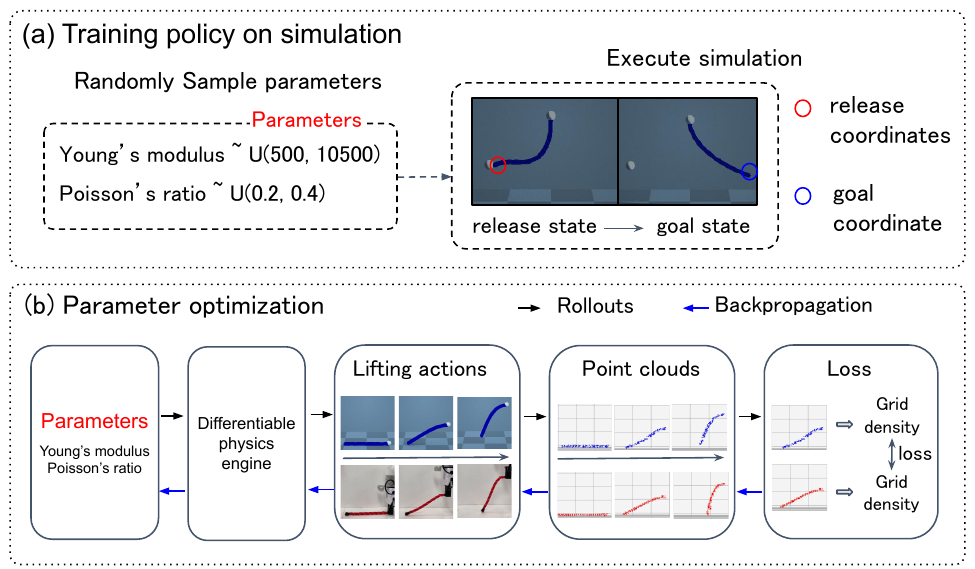}
    \vskip -0.05in
    \caption{Overview of our Pipeline: (a) We randomly set parameters in a simulation to collect release and goal coordinates. Using this dataset, we trained a simple MLP policy. (b) For inference, parameters are optimized from real-world objects by comparing point cloud grid densities between simulation and reality through differentiable physics.}
    \label{fig:real2sim2real}
    \end{center}
    \vskip -0.3in
\end{figure}

To achieve generalizable deformable manipulation, we propose GenDOM, which consists of two primal parts: (1) \emph{training augmented policy with diverse parameters on simulator} and (2) \emph{estimating deformable object parameters during the inference}. The entire GenDOM pipeline is illustrated in Figure~\ref{fig:real2sim2real}. We will first provide the task description, followed by a detailed presentation of our framework.

\subsection{{Task Description}}
In this paper, we propose a generalizable framework for manipulating a diverse range of deformable objects.
To evaluate the generalizability of our framework, we utilize different deformable objects, specifically ropes and cloths.
We consider three tasks showcasing dynamic deformation, including Rope reaching, Rope casting, and Cloth spreading, as depicted in Figure~\ref{fig:tasks}. Due to the space limitation, we use Rope reaching as an example to explain the problem setup. In this task, we are given a goal coordinate $g$ and aim to train a manipulation policy that can control a robot arm to handle the end-point of this rope and move it to a releasing coordinate $r$ in mid-air so that
the rope can move to the given goal coordinate $g$ under the influence of gravity during its initial swing once released from $r$. 
The other types of deformable object manipulation are given in Section \ref{sec:task}.

\subsection{Training Policy with Diverse Parameters in Simulator} \label{sec:policy}

The main challenge for learning a universally applicable deformable object manipulation policy arises due to the variation in physical mechanics across objects. 
These attributional differences significantly influence the dynamics and the optimal robot control.
To address this issue and learn such a generalizable deformable object manipulation policy, our key idea is to augment the policy condition with deformable object parameters. 
This augmentation enables the policy to be aware of the dynamic variations and adjust its actions accordingly for objects with different dynamics, thus enhancing its generalization ability for deformable object manipulation. Specifically, we propose to further condition policy on Young’s modulus and Poisson’s ratio~\cite{sanchez2018robotic}. 
These two factors serve as critical determinants of a deformable object's behavior, representing stiffness and deformation characteristics, respectively. 
Both factors are frequently employed in the modeling of deformation \cite{sengupta2020simultaneous, dom-pt1}, and our paper is the first work to condition manipulation policy $\pi$ on these two factors to achieve generalizable deformable object manipulation.
As iteration cost in simulated environments is low, we opt to train our parameter-aware policy on simulation first. 
For each demonstration, we randomly sample Young’s modulus parameter $p^y$ and Poisson’s ratio $p^l$.
In this study, we utilize imitation learning for policy formulation as below.
\begin{equation}
\vspace{-0.15em}
\mathcal{L}_{\theta}\ =\  \frac{1}{N}\sum_{i=1}^{N} (  X - \pi_{\theta}(Y,  p^y_{i}, p^l_{i}))^2
\vspace{-0.15em}
\end{equation}
where $\theta$ is the trainable parameter of the manipulation policy $\pi$, and $N$ is batch size.
Depending on the desired task, by designing the policy's output $X$ and input $Y$, \emph{e.g.}, the release coordinate is $X$, and the goal coordinate is $Y$ in Rope reaching task, the proposed framework can adapt to various deformable objects and various tasks.



\subsection{Gradient-based Real2Sim Parameter Optimization}

After obtaining our augmented manipulation policy trained across an assortment of deformable object parameters, it remains challenging to apply this policy to real-world tasks due to the unavailability of specific object parameters during real-world deployment. To mitigate this obstacle, we introduce the Real2Sim method, which estimates these parameters using a single real-world demonstration.
We begin by transforming a single real-world demonstration of a deformable object into point clouds, taking advantage of the inherent versatility of point clouds in representing diverse shapes and accommodating significant deformations, which makes them an ideal choice for simulating deformable objects. 
Utilizing differentiable physics~\cite{hu2019difftaichi,huang2021plasticinelab}, we adjust and estimate the deformable object parameters in the simulation to minimize the divergence between the grid density of point clouds from real-world demonstrations and simulations. 
As depicted in Figure~\ref{fig:real2sim2real} (b), our system is constructed on the framework of PlasticineLab~\cite{huang2021plasticinelab}, where a differentiable simulator enables gradient-based trajectory optimization through particle dynamics.
We perform predefined actions from the same initial states in both the simulated and real-world environments; the robot vertically lifts the object at a constant speed from either the end of the rope or the corner of the cloth.
To capture the real-world demonstration, we employ two RGB-D cameras to record the trajectory and preserve its point clouds. At the same time, we also preserve the point clouds of the trajectory in simulation.
However, due to the difference in the sequence and number of points between the simulator and real-world environments, directly calculating the loss between point clouds of simulation and real-world demonstration is infeasible. 

To bypass this hurdle, we represent the point clouds from both simulation and real-world demonstration into a 3-dimensional grid density space (discretized into 64 $\times$ 64 $\times$ 64 in our study) to synchronize them, regardless of their differing quantities and orders. Explicitly, each cloud's position is computed within the grid, assigning a weight to each grid cell based on its fractional position $f_x$ for superior density matching. Particularly, these weight assignments are attenuated based on cloud's distance to the cell's center, where the closer to the center, the higher the weight, thereby promoting a smooth distribution of the particle's mass across neighboring grid cells. This smooth distribution of the particle's mass enables successful alignment between the real-world and simulated scenarios, thereby utilizing the prior knowledge in the differentiable physics simulation to ensure an accurate loss that converges even with a single demonstration. By minimizing the L1 loss between the grid densities of the simulator and the real-world environments, with Young’s modulus and Poisson’s ratio serving as trainable variables, we can indirectly estimate these parameters through the differential simulator. We then condition these parameters directly on the pre-trained manipulation policy $\pi$ in Section \ref{sec:policy} to execute the manipulation task. 
\vspace{-0.5em}
\section{Experiments}
\vspace{-0.5em}
\label{sec:experiments}
\begin{figure}[t]
    \begin{center}
    \includegraphics[width=0.5\textwidth]{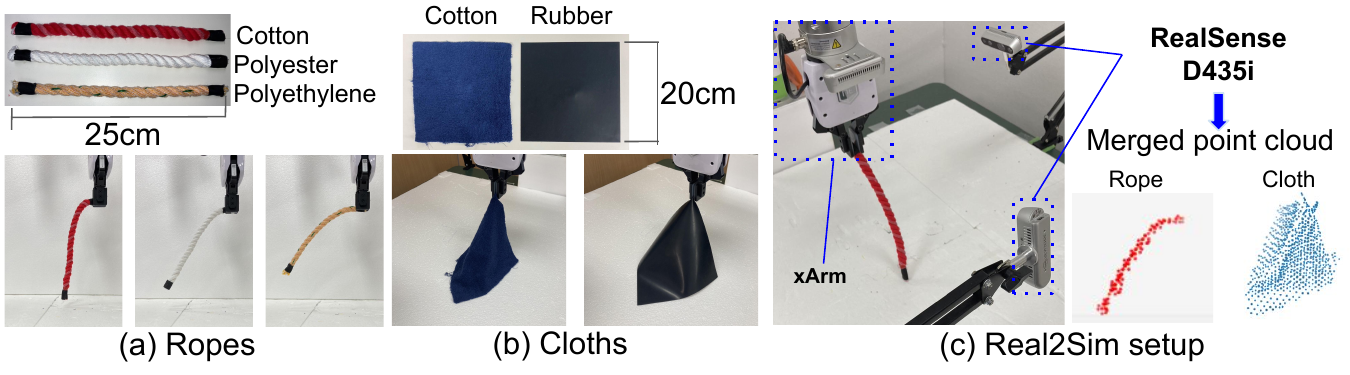}
    \vskip -0.1in
    \caption{(a) Three different type ropes: cotton rope, polyester rope, and polyethylene rope. (b) Two different types of cloth: cotton cloth and rubber cloth. (c) Our experiments setup for Real2Sim: two Intel Realsense Depth Camera D435i and xArm 7. 
    }
    \label{fig:setup}
    \end{center}
   \vskip -0.3in
\end{figure}

In this section, we experimentally evaluate our method in both simulation and real world. Our experiments seek to address the following questions:
\begin{itemize} 
    \item Can our gradient-based optimization using a designed loss function estimate the physical parameters of ropes (Young’s Modulus and Poisson’s ratio) given a limited demonstration? Is it better than a learning-based method? (\ref{subsec:sim2sim} and~\ref{subsec:real2sim}) 
    \item Does the estimated physical parameters help our parameter-aware policy to generalize diverse setups (in-domain, out-of-domain, and real-world deployment)? (\ref{subsec:ablation}, \ref{subsec:generalization_test}, and \ref{subsec:real_robot}) 
    \item Can our gradient-based optimization method and parameter-aware policy be deployed for cloth? (\ref{subsec:extension}) 
    \vspace{-0.1em}
\end{itemize} 



\subsection{Tasks} \label{sec:task}
As Figure~\ref{fig:setup} (a) and (b) show, we use three different types of ropes: cotton, polyester and polyethylene ropes, and two different cloths: cotton and rubber cloths to evaluate the effectiveness of our method on different deformable objects. 
Furthermore, to demonstrate our framework's capability to handle deformable object manipulation with varying dynamics, we evaluated it on the following  tasks:

\textbf{Rope reaching:} We randomly sample release coordinate $r$ with the parameters.
Given the sampled rope parameters, we record the maximum reach distance of the bottom tip of the rope as the goal coordinate $g$. Utilizing $g$ as hindsight for $Y$, the policy is trained to predict $r$ for $X$.

\textbf{Rope casting:}  Given a goal coordinate $(g_x, g_y)$, a manipulation policy is trained to operate a robot arm to handle the top end of this rope and take one momentary action $a$ in mid-air. 
The rope is expected to move towards $(g_x, g_y)$ during its initial swing.
We randomly sampled an action $a$ that moves instantaneously and horizontally relative to the ground with the parameters.
Given the sampled rope parameters, we record the highest reach coordinate of the bottom tip of the rope as the goal coordinate $(g_x, g_y)$. Utilizing $(g_x, g_y)$ as hindsight for $Y$, we train our policy $\pi$ to predict $a$ for $X$.

\textbf{Cloth spreading:}  The objective is to spread a cloth over a table. 
A manipulation policy is trained to control a robot arm to handle the center of the cloth and take actions with acceleration in mid-air. 
The robot moves 16cm in 16 steps. 
For each step $i$, the distance traveled for each step is defined as: 
$1 - (4.5 - 0.5i) \times \text{$A$},  \text{for } i = 1, \dots, 8$ and $1 + (0.5i - 4.0) \times \text{$A$},  \text{for } i = 9, \dots, 16$.
If $A$ is large, the robot's movement range shifts from small to large, resulting in a greater acceleration.
We randomly sampled $A$ and cloth parameters, and recorded the minimum value of $A$ that resulted in a spread of at least 75\% of the cloth area being smooth and wrinkle-free.
The policy is trained to predict $A$ as $X$ with the parameters. $Y$ is unset for this task.

\begin{table*}[!htb]
\centering
\footnotesize
\setlength{\tabcolsep}{0.1mm}
\renewcommand{\arraystretch}{1.2}
\setlength{\tabcolsep}{1.5pt}
\caption{
Summary of our methods and baselines in deformable object manipulation
}
\vskip -0.1in
\begin{tabular}{@{}rc|cc|cccc@{}}
\toprule
\multicolumn{2}{c|}{\emph{Category \& Method}}                                            & \multicolumn{2}{c|}{\emph{Policy}}     & \multicolumn{4}{c}{\emph{Adaptation}}                 \\ 
Category           & Method                                                               & Parameter-aware    & Data              & Type       & Speed  & \#demo in real & S2R gap \\ \midrule
Real               & Many \cite{nair2017combining,lee2021sample,shivakumar2023sgtm}       & \ding{55}          & Real (single)     & None       & N/A    & Many           & None    \\
Real               & Many \cite{yan2021learning, yan2020self}                             & \ding{55}          & Real (diverse)    & None       & N/A    & Many           & None    \\
Sim2Real           & Many \cite{matas2018sim, wu2019learning,lin2021softgym}              & \ding{55}          & Sim (diverse)     & None       & N/A    & 0 $\sim$       & High    \\
Real2SimReal       & R2S2R~\cite{lim2022real2sim2real}                                    & \ding{55}          & Sim (Single)      & Train      & Low    & 20 $\sim$      & Low     \\
Real2SimReal       & GenDOM (Ours)                                                        & \ding{51}          & Sim (diverse)     & Condition  & High   & 1 $\sim$       & Low     \\ \bottomrule
\end{tabular}\label{tbl:correspondence}
\vskip -0.15in
\end{table*}

\subsection{Environments Setup}
\label{subsec:setup}

\textbf{Deformable Parameter Optimization:}
To bridge the gap between the behavior of deformable objects in the real world and in simulation, we use gradient-based optimization to estimate both Young's modulus and Poisson's ratio.
Our hardware configuration, depicted in Figure~\ref{fig:setup} (c), incorporates two Intel Realsense Depth Camera D435i to obtain point clouds and xArm 7 to manipulate rope and cloth. 
Two camera poses are calibrated to the xArm-world coordinate. 
We take a demonstration for each object by executing the predefined lifting actions. Both real-world and simulation point clouds are transferred into a 3-dimensional grid density of size (64, 64, 64).
We use PointNet++~\cite{qi2017pointnet++} \textbf{as the baseline}. 
We adopt PointNet++ for its proficiency in processing point clouds, known for its versatility in representing diverse shapes and handling deformations.
We randomly assign parameters to the simulation and execute the predefined actions to obtain the last state of either the rope or cloth within the simulation.
Using the 998 sets of parameters and the last state, PointNet++ is trained to output the parameters from the last state. The results analysis are given in Section \ref{subsec:sim2sim} and \ref{subsec:real2sim}.

                   

\textbf{Policy Training:} 
We train our policy on a synthesized dataset that covers diverse physical parameters for deformable objects. Specifically, we randomize Young’s modulus and Poisson’s ratio within the ranges [1500, 8200] and [0.34, 0.36] for the simulation test, and we randomize them in [500, 10500] and [0.2, 0.4] for the real world test.  
We use the recent Real2Sim2Real method \textbf{(R2S2R)} ~\cite{lim2022real2sim2real} as the baseline where
it estimates parameters and executes a sim2real transfer with a policy trained by the dataset collected with estimated parameters. 
We follow R2S2R's setting and prepare two datasets to train it in simulation and the real world, respectively.
Two datasets are (1) A dataset covering all three rope parameters, and (2) a dataset only covering the red rope parameter. 
\textbf{We collect 2,271 demonstrations for simulation evaluation and 40,000 for real-world evaluation for all datasets to ensure a fair comparison between the proposed method and baselines.}
Furthermore, in the comparison in the real world, we prepare a policy trained on only the real-world dataset ($\pi_{RD}$): (1)  300 demonstrations for three ropes together and (2) 100 demonstrations with only the red rope's parameters. 
Initializing policy weights with those trained in simulation,  we also developed an adapted policy by fine-tuning it using 300 real-world demonstrations. 
The results analysis are given in Section \ref{subsec:generalization_test}. We summarize the baseline methods and the relevant information in Table~\ref{tbl:correspondence}.

\begin{table}[!t]
\centering
\renewcommand{\arraystretch}{1.2}
\setlength{\tabcolsep}{1.5pt}
\caption{
Comparison between our gradient-based method and the baseline (PointNet++) regarding the difference between estimated and ground-truth parameters and the Chamfer distance within a Sim2Sim setup.
}
\vskip -0.05in
\begin{tabular}{cccc}
\hline
                   & \textbf{Young's modulus}  & \textbf{Poisson's ratio}      & \textbf{Chamfer}                \\
\textbf{Method}    & Mean of diff              & Mean of diff                  & \textbf{Distance}               \\ \hline
Ours (all steps)   & $\mathbf{11.69}\pm 8.28$  & $\mathbf{0.0032}\pm 0.0074$   & $ \mathbf{0.0004}\pm 0.0005$    \\
Ours  (last step)  & $13.74\pm 6.75$           & $0.0067\pm 0.0098$            & $0.0004\pm 0.0004$              \\ 
Pointnet++         & $173.51\pm 200.14$        & $0.029\pm 0.024 $             & $0.0025\pm 0.0016$              \\ \hline
\end{tabular}\label{tbl:sim2sim}
\vskip -0.25in
\end{table}

\subsection{Sim2Sim Performance}
\label{subsec:sim2sim}
To verify the performance of the proposed method in estimating the ground truth Young's modulus and Poisson's ratio parameters, we conducted evaluations on the simulation.
Table~\ref{tbl:sim2sim} compares our gradient-based and learning-based methods (Pointnet++) on the accuracy of sim2sim parameter optimization for rope. 
We sampled five sets of ground-truth parameters and reported the mean and standard deviation. 
In addition, we calculate the Chamfer distance between point clouds of the last state with ground-truth parameters and estimated parameters.  
For the gradient-based method, we sample parameters as an initial state for each set of ground-truth parameters and proceed with optimization. 
PointNet++ estimates parameters from each last state after executing the predefined actions. 
We also test gradient-based optimization only with the last state loss for a fair comparison. 
From the results, the gradient-based method yields a \textbf{significantly smaller} mean and standard deviation than PointNet++. 
When comparing the gradient-based method, the gradient-based with the last step provides nearly identical performance with the gradient-based with all steps.
We utilize the gradient-based with all step loss in subsequent experiments since it shows slightly more accuracy.

\subsection{Real2Sim Performance}
\label{subsec:real2sim}

To assess the efficacy of our proposed parameter estimation technique in the real world, we conducted an evaluation by estimating parameters from a single rope demonstration.
Table~\ref{tbl:real2sim} shows the estimated parameters and Chamfer distance between the last state of the real world and the simulation with estimated parameters. 
We utilize Chamfer distance since real-world ground truths aren't accessible.
The gradient-based method optimizes parameters from five randomly sampled parameters as the initial state for each object. 
Pointnet++ uses point clouds of the last state to estimate parameters. 
For all ropes, Chamfer distance of the gradient-based method \textbf{shows a smaller value} than PointNet++.
The estimated Young’s modulus and Poisson’s ratio are [1779.38, 0.35], [3276.12, 0.346], and [8000.31, 0.36] for cotton, polyester, and polyethylene ropes, respectively.
Our method can directly estimate parameters \textbf{without time-consuming data collection.}
A supplemental video showcases trajectory comparisons between real and simulated objects with estimated parameters. 

\begin{table}[!t]
\centering
\footnotesize
\setlength{\tabcolsep}{1.5pt}
\renewcommand{\arraystretch}{1.2}
\caption{Comparison of Real2Sim rope parameter optimizations on the optimized parameters and Chamfer distance.}
\vskip -0.1in
\begin{tabular}{ccccc}
\hline
                 &                  & \textbf{Young's}             & \textbf{Poisson's}           & \textbf{Chamfer}               \\
\textbf{Rope}    &\textbf{Method}   & \textbf{modulus}             & \textbf{ratio}               & \textbf{Distance}              \\ \hline
Cotton           & Ours   & $ \mathbf{1779.38}\pm 7.24$  & $ \mathbf{0.35}\pm 0.04$     & $\mathbf{0.034}\pm 0.00$       \\ 
                 & Pointnet++       & $509.83$                     & $0.30$                       & $0.092$                        \\ \cline{1-5}
Polyester        & Ours  & $\mathbf{3276.12}\pm 3.71$   & $ \mathbf{0.346}\pm 0.03$    & $\mathbf{0.060}\pm 0.00$       \\ 
                 & Pointnet++       & $4660.12$                    & $0.30$                       & $0.067$                        \\ \cline{1-5}
Polyethy-        & Ours   & $\mathbf{8000.31}\pm 48.04 $ & $ \mathbf{0.36}\pm 0.00$     & $\mathbf{0.053}\pm 0.00$       \\
lene             & Pointnet++       & $7305.96$                    & $0.30$                       & $0.057$                        \\ \hline
\end{tabular}\label{tbl:real2sim}
\vskip -0.25in
\end{table}

\subsection{Ablation Study}
\label{subsec:ablation}
To discern the contribution of each parameter to the tasks, we evaluate the accuracy by modulating the inputs to the network for rope reaching and rope casting in simulation. 
This assessment is conducted with four distinct input configurations: both Young's modulus and Poisson's ratio, only Young's modulus, only Poisson's ratio, and neither of these parameters.
The policy for each task is trained with 40,000 and 20,000 demonstrations, respectively. The parameters are in the range of [500, 10,500] for Young’s modulus and [0.2, 0.4] for Poisson’s ratio.
We randomly sample 30 new goal coordinates and parameters. 
Table~\ref{tbl:ablation} shows the best performance for the policy trained on both parameters, suggesting that the \textbf{two parameters in conjunction are informative} for the tasks. 

\begin{table}[!t]
\centering
\renewcommand{\arraystretch}{1.2}
\caption{Ablation study results. For training, we utilize four variations of inputs. We report the mean and standard deviation on 30 new goal coordinates. 
}
\vskip -0.05in
\begin{tabular}{ccccc}
\hline
                & \textbf{Both}      & \textbf{Young's}  & \textbf{Poisson's}  & \textbf{No}           \\
                & \textbf{parameters}& \textbf{modulus}  & \textbf{ratio}      & \textbf{parameters}   \\ \hline
Rope            & $ \mathbf{0.024}$  & $0.034$           & $0.062$             & $0.062$               \\
reaching        & $ \pm 0.021$       & $\pm 0.040$       & $\pm 0.058 $        & $\pm 0.059 $          \\ \hline
Rope            & $ \mathbf{0.065}$  & $0.067$           & $0.077$             & $0.076$               \\
casting         & $ \pm 0.031$       & $\pm 0.033$       & $\pm 0.041 $        & $\pm 0.042 $          \\ \hline
\end{tabular}\label{tbl:ablation}
\vskip -0.1in
\end{table}

\subsection{Generalization Test}
\label{subsec:generalization_test}
To test generalizability to various ropes, we evaluate the policy on in-domain (ID) and out-of-distribution (OOD) parameters in simulation. For the ID test, we randomly sample 40 new goal coordinates with parameters of three ropes. 
For the OOD test, we randomly sample 40 new goal coordinates with parameters range [500, 1500] and [0.3, 0.33] for Young’s modulus and Poisson’s ratio, respectively. 
Table~\ref{tbl:id_ood} summarizes our findings. 
(1) Our method performs 62{\%} better than R2S2R (trained on three ropes) for ID. 
(2) Our method performs well even in the OOD setup. Specifically, we compare our method with R2S2R (trained on the red rope) for OOD. 
Even though the baseline is trained on the red rope's parameter, which is close to the OOD parameters' range, our method performs 15{\%} better than R2S2R (trained on the red rope). 
This suggests our parameter-conditioned policy has a \textbf{stronger generalization ability} against variable deformable object dynamics.

\begin{table}[!t]
\centering
\renewcommand{\arraystretch}{1.2}
\caption{Comparison of generalization for ID and OOD rope parameters (Calculated on new 40 goal coordinates). 
}
\vskip -0.05in
\begin{tabular}{ccc}
\hline
\textbf{Method}                 & \textbf{ID}                   & \textbf{OOD}                     \\ \hline
Ours                            & $ \mathbf{0.031 \pm 0.050}$   & $ \mathbf{0.039 \pm 0.042}$      \\ 
R2S2R (Three ropes)             & $0.082 \pm 0.081$             & $0.096 \pm 0.104$                \\ 
R2S2R (Cotton rope)             & $0.047 \pm 0.033$             & $0.046 \pm 0.044$                \\ \hline
\end{tabular}\label{tbl:id_ood}
\vskip -0.22in
\end{table}


\subsection{Real Robot Experiments}

\label{subsec:real_robot}
Finally, we deploy our parameter-aware policy in the real world using the estimated parameters for each rope. We set a goal conditioned from [27.5, 30.0, 32.5, 35, 37.5, 40.0, 42.5] (cm), as represented by the range depicted in Fig~\ref{fig:tasks}, and tested the policy three times for each goal coordinate. We sum up the difference for each goal point and calculate its mean and standard deviation on three times trials, with results in Table~\ref{tbl:sim2real}. For red rope, our method without adaptation performs 10{\%} better than $\pi_{RD}$ (trained on the red rope). This shows that our method, only using a single real-world demonstration, achieves better performance than the real-world policy trained with 100 demonstrations. With adaptation, our method (sim+real) performs 32{\%} better than $\pi_{RD}$ and 44{\%} better than R2S2R (trained on the red rope).

\begin{table}[t]
\centering
\setlength{\tabcolsep}{1pt}
\renewcommand{\arraystretch}{1.1}
\caption{Results of real-world rope reaching, summarizing the differences for goal coordinates with the mean and standard deviation from three trials.}
\vskip -0.05in
\begin{tabular}{ccccccc}
\hline
                   & \textbf{Ours}          & \textbf{Ours}        & {$\mathbf{\pi_{RD}}$}  & {$\mathbf{\pi_{RD}}$}& \textbf{R2S2R}    & \textbf{R2S2R}    \\
\textbf{Rope}      & sim+                   & no                   & Three                  & Cotton               & Three             & Cotton     \\ 
\textbf{reaching}  & real                   & adaptation           & ropes                  & rope                 & ropes             & rope     \\ \hline
Cotton             & $ \mathbf{0.112}$      & $ \mathbf{0.148}$    & $0.426$                & $0.164$              & $0.318$           & $0.200$           \\ 
                   & $ 0.007$               & $ \pm 0.012$         & $\pm 0.017$            & $\pm 0.012$          & $\pm 0.004$       & $\pm 0.021$       \\ \hline
Polyester          & $ \mathbf{0.076}$      & $ \mathbf{0.106}$    & $0.273$                & $0.512$              & $0.343$           & $0.464$           \\
                   & $ \pm 0.025$           & $\pm 0.005$          & $\pm 0.056$            & $\pm 0.039$          & $\pm 0.017$       & $\pm 0.038$       \\ \hline
Polyethylene       & $ \mathbf{0.193}$      & $0.588$              & $0.297$                & $0.401$              & $0.364$           & $0.473$           \\ 
                   & $ \pm 0.009$           & $\pm 0.070$          & $\pm 0.113$            & $\pm 0.060$          & $\pm 0.061$       & $\pm 0.024$       \\ \hline
\end{tabular}\label{tbl:sim2real}
\vskip -0.1in
\end{table}

\begin{table}[t]
\renewcommand{\arraystretch}{1.0}
\setlength{\tabcolsep}{1pt}
\centering
\caption{Real-world Cloth spread results with mean and standard deviation on three times trials. Additionally, we provided the proportion in relation to $A$ without parameters as acceleration ratio.}
\vskip -0.05in
\begin{tabular}{ccccc}
\hline
                              & \multicolumn{2}{c}{\textbf{Both parameters}}        & \multicolumn{2}{c}{\textbf{No parameters}}       \\ 
\textbf{Cloth}                & \textbf{Extended}           & \textbf{Acceleration} & \textbf{Extended}     & \textbf{Acceleration}    \\ 
\textbf{spreading}            & \textbf{ratio}              & \textbf{ratio}        & \textbf{ratio}        & \textbf{ratio}           \\ \hline
Cotton                        & $\mathbf{0.78} \pm 0.08$    & $\mathbf{1.70}$       & $0.52 \pm 0.03$       & $ 1.0 $                  \\ 
Rubber                        & $ 0.98 \pm 0.01$            & $\mathbf{0.26}$       & $ 0.97 \pm 0.01$      & $ 1.0 $                  \\ \hline
\end{tabular}\label{tbl:sim2real_cloth}
\vskip -0.2in
\end{table}


\subsection{Extension to Cloth}
To demonstrate the versatility of our proposed method for other deformable objects, we conducted parameter optimization and real-world deployment tests using cloth.

\label{subsec:extension}
\textbf{Parameter optimization} We use the same predefined action and setup for the rope parameter optimization. The robot grasps one of the four corners of the cloth on the desk and lifts it vertically at a constant speed. The gradient-based method optimizes parameters from five randomly sampled parameters as the initial state for each cloth. The estimated Young’s modulus and Poisson’s ratio are [584.21, 0.25] and [6828.33, 0.27] for cotton and rubber cloth, respectively.


\textbf{Real-world deployment} 
To evaluate the contribution of the parameters, we compare the accuracy by modulating the inputs to the network for cloth spreading task in the real world. 
We train our policy using an 800 dataset with Young’s modulus and Poisson’s ratio within the ranges [500, 10500] and [0.2, 0.4]. 
In addition, we fine-tuned the policy using two demonstrations for two cloths.
This assessment is conducted with the policy trained with both Young's modulus and Poisson's ratio and one by neither of these parameters.
We conducted five trials for each type of cloth and evaluated how much it could spread in its expanded state.
In addition, based on the $A$ by the policy without parameters, we normalized other $A$ as the acceleration ratio.
The results are presented in Table~\ref{tbl:sim2real_cloth}. 
Without parameters, the same acceleration is output for both cotton and rubber cloths. 
When using the policy with parameters, it enhances the spreading of cotton by 50\%, applying higher acceleration. In contrast, for rubber, the policy achieves spreading with an acceleration reduced by 64\%.
To properly execute the task, it can adjust the speed, and it also suppresses risks such as malfunctions.


\section{Conclusion}
In this paper, we present a framework designed for one-shot manipulation of deformable objects, specifically ropes and cloths. By combining a pre-trained, parameter-conditioned policy with a gradient-based parameter optimization technique, we show that our method outperforms the baselines in both in-domain and out-of-distribution settings across simulation and real-world rope manipulation tasks, as well as in real-world deployment for both ropes and cloths.

\section*{Acknowledgement}
This research was supported in part by KIOXIA Corporation.

\clearpage

\printbibliography
\clearpage

\end{document}